\documentclass[10pt,twocolumn,letterpaper]{article}

\usepackage{cvpr}
\usepackage{times}
\usepackage{epsfig}
\usepackage{graphicx}
\usepackage{amsmath}
\usepackage{amssymb}
\usepackage{subfigure}
\usepackage{color}
\usepackage{setspace}
\usepackage{algorithm}
\usepackage{algorithmic}
\usepackage{multirow}

\usepackage[pagebackref=true,breaklinks=true,letterpaper=true,colorlinks,bookmarks=false]{hyperref}

\cvprfinalcopy 

\def\cvprPaperID{1736} 

\ifcvprfinal\fi

\begin{document}
\title{Learning from Noisy Labels with Noise Modeling Network}

\author{Zhuolin Jiang,  Jan Silovsky\thanks{Now at Apple, Cambridge, MA.},  Man-Hung Siu\footnotemark[1],  William Hartmann,  Herbert Gish,  Sancar Adali \\
Raytheon BBN Technologies \\
10 Moulton Street, Cambridge, MA 02138 \\
}

\maketitle

\begin{abstract}
Multi-label image classification has generated significant interest in recent years and the performance of such systems often suffers from the not so infrequent occurrence of incorrect or missing labels in the training data. In this paper we extend the state-of the-art of training classifiers to jointly deal with both forms of errorful data. We accomplish this by modeling noisy and missing labels in multi-label images with a new Noise Modeling Network (NMN) that follows our convolutional neural network (CNN), integrates with it, forming an end-to-end deep learning system, which can jointly learn the noise distribution and CNN parameters. The NMN learns the distribution of noise patterns directly from the noisy data without the need for any clean training data. The NMN can model label noise that depends only on the true label or is also dependent on the image features. We show that the integrated NMN/CNN learning system consistently improves the classification performance, for different levels of label noise, on the MSR-COCO dataset and MSR-VTT dataset. We also show that noise performance improvements are obtained when multiple instance learning methods are used.   
\end{abstract}

\section{Introduction}
Deep convolutional neural networks (ConvNet) have shown impressive performance in various vision tasks (\textit{e.g.} object recognition~\cite{He16} and detection~\cite{He17}). However, these achievements often require large amounts of training data with unambiguous and accurate annotations, such as ImageNet~\cite{Deng09}. Relying on humans for annotating images to create such datasets is prohibitively expensive and time-consuming. To relax this limitation, some training paradigms which aim to reduce the need for expensive annotation have been developed, such as unsupervised learning~\cite{Le12}, weakly supervised learning~\cite{Joulin16,Su16} and self-supervised learning algorithms~\cite{Pinto16,Wang15}.

An alternative approach is to use labels from readily-available sources of annotated data, such as user tags, captions from social networks, or keywords from image search engines. The labels in these datasets are noisy and unreliable, adversely affecting the model learning process. Because of the readily available nature of noisily labeled datasets, various approaches have been proposed to handle model training with noisy labels. A simple approach is to remove those suspect or noisy examples from the dataset or re-label them by expert labellers~\cite{Barandela00,Brodley99,Reed14,Liu16}. In addition to the difficulty of distinguishing between mislabeled samples and hard-to-label samples, this could still involve substantial effort by expert labellers. Recent efforts have been focused on building robust neural network models that can be trained using data with unreliable and noisy labels~\cite{Bekker16,Xiao15,Goldberger17,Patrini17,Mnih12,Veit17}. In \cite{Xiao15,Bekker16,Mnih12},  EM-based algorithms were developed where the true labels are hidden and estimated in the E-step. A neural network is then retrained in the M-step using the estimated true labels. The idea of using latent true labels for model estimation could be powerful.  However, iterating between EM-steps and network retraining does not scale well because even one training iteration is non-trivial for large neural networks. Extending the hidden label framework,~\cite{Patrini17} recently introduced two loss correction approaches to handle label noise in multiclass image classification problems using label transition matrices separately trained from a dataset with clean labels. In~\cite{Patrini17}, the noise distribution depends on the hidden true label only, without considering input features.  Extending~\cite{Patrini17}, \cite{Goldberger17} used an additional softmax layer to generate a transition matrix that is dependent on both the input features and the true labels.

\begin{figure*}[t]
        \begin{center}
                \includegraphics[width=0.68\linewidth]{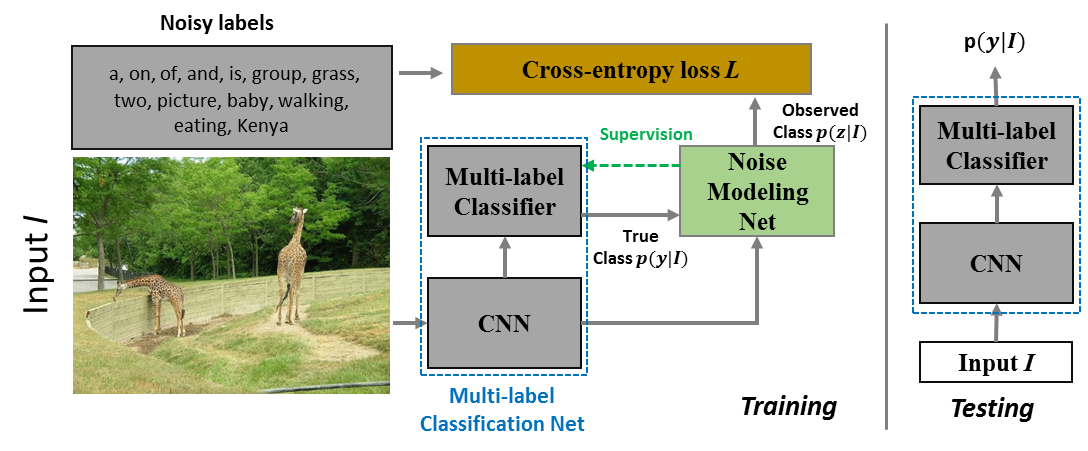}
        \end{center}
        \caption{High-level overview of our noise modeling neural network architecture for the training phase (left) and test phase (right). Noisy labels are used as targets to train the noise modeling net. The knowledge of noise modeling net is transferred to the final multi-label classifier, by treating its inferred posterior probabilities of hidden true label as `soft target' labels for training the classifier. The noise modeling network and multi-label classification network are trained simultaneously.}
        \label{overview1}
\end{figure*}

In this paper, we explore how to effectively learn discriminative models from noisy data for multi-label classification where each image can be tagged with one or more labels. The labels for an image may or may not be correlated with one another. For example, in Figure~\ref{overview1}, an image of giraffes eating at a grassland may include the label ``giraffe'', ``eating'', ``green'' and ``grass''. ``Grass'' and ``green'' may be highly correlated; often, ``grass'' is implicitly assumed to be green and ``green'' may not be tagged, introducing labeling noise. Instead of simply modeling flipping of labels from one class to another in multi-class classification, there are two types of label noise in multi-label classification, namely, missing labels in which a true label is missed in the annotation, or incorrect labels in which a label is erroneously marked. We followed the hidden label framework with a multi-label classifiction network (MLCN) to estimate the hidden true label and introduced the noise modeling network (NMN) as shown in Figure~\ref{overview1}. The noise modeling network captures the feature patterns of both types of noise and learns the noise label distribution which is undesirable during inference.  That is, the goal is to build the best classifier for predicting the ``true label'' even when these  ``true'' labels are not explicitly observed in training.  Therefore, during test, only the MLCN is used to infer clean image labels. Similar to the other hidden true label approaches, the noise modeling network implicitly provides supervision---the posterior probability of hidden true labels---to guide the multi-label classifier training. The combination of the MLCN and NMN allows our model to learn the noise distribution, the posterior of hidden true label, and classifier network parameters jointly. This is in contrast to previous approaches, which iteratively solve two sub-problems with EM based algorithms to approximate a joint solution~\cite{Xiao15,Bekker16,Mnih12}, or learn the noise distribution from a separate validation set~\cite{Patrini17}. Unlike recent approaches~\cite{Veit17,Li17} that require a separate clean labeled dataset to bootstrap parameter estimation, our approach is completely trained on a noisy dataset. The contributions of our paper can be summarized as:

\vspace{-0.2cm}
\begin{itemize}
\item We introduce an auxiliary noise modeling network to cope with label noise in training of a multi-label classification system based on neural networks. NMN can be integrated into different types of neural networks.
\item The noise model used by NMN can cope with both incorrect and missing labels.
\item We compare feature-independent and feature-dependent noise modeling and demonstrate superiority of the latter.
\item We show that training with the auxiliary NMN and the EM-based approach---applied in earlier work to tackle noisy labels---are analogous. The iterative nature of EM-based approach makes it intractable for larger datasets. In our approach, both the MLCN and noise models are fitted simultaneously during the Stochastic Gradient Descent (SGD) training, leading to earlier convergence. Effectively, through back-propagation, the NMN infers the posterior of the hidden true label as soft targets in training of the MLCN. This resembles the student-teacher concept where the NMN acts as teacher and the MLCN as student.
\item Multiple Instance Learning (MIL) has been shown to improve performance in the task of concept detection. We demonstrate that the technique also improves robustness against different types of label noise in training and show our NMN augmented training and MIL are complementary.
\end{itemize}

\subsection{Related work}

Learning from noisy labels is an active research topic in the field of machine learning.
Current learning approaches can be roughly divided into two categories. The first class of approaches directly learns from data with noisy labels by developing noise robust algorithms~\cite{Freney14,Kaban12,Natarajan13,Reed14,Liu16}. The effect of label noise in model training has been well studied in logistic regression and SVMs~\cite{Freney14,Kaban12}.~\cite{Natarajan13} developed methods for suitably modifying any surrogate loss function and showed that minimizing the modified loss function on noisy data can provide a performance bound.~\cite{Reed14} changed the targets label on the fly, in order to reduce the damage from the noisy samples.~\cite{Liu16} proposed to compute importance weights for training samples and combined them into the loss functions. The second class of approaches are semi-supervised~\cite{Grandvalet05,Lee13}, which assumes some fraction of data has clean labels, while the rest are either unlabeled or have noisy labels.~\cite{Grandvalet05} added entropy maximization regularization to the unlabeled data to encourage their class predictions with high confidence.

There are limited studies which explicitly address the problem of training a neural network from noisy labels. Most of existing work follows the hidden label framework, but focuses on the multi-class classification task~\cite{Larsen98,Mnih12,Sukhbaatar15,Xiao15,Goldberger17,Patrini17,Li17,Guan18}.~\cite{Larsen98} models symmetric label noise (\textit{i.e.} independent of the true label).~\cite{Sukhbaatar15,Guan18,Bekker16,Patrini17} model asymmetric label noises that are dependent on the true label, but independent of the input features.~\cite{Xiao15} developed an image-dependent noise model to predict label noise types but their model requires access to a small clean dataset to estimate the label transition matrix.~\cite{Goldberger17} introduced a feature-dependent noise adaptation layer on top of the softmax to absorb label noise, but it is designed for a single label per image (\textit{i.e.} multi-class classification).~\cite{Veit17} presents a label cleaning network, which maps noisy labels to clean labels, and the network is dependent on input features. This approach requires a small fraction of images with clean annotations. The closest work is from~\cite{Misra16} which proposed a similar noise model for missing labels in multi-label classification, and showed that the complex, human-centric prediction model is better than the simple, visual presence prediction model in missing label scenarios. However, this human-centric prediction model may not work in incorrect label scenarios. In addition, it focuses on improving inference in generating human-centric labels which continues to include labeling noise, instead of predicting clean labels.

Compared to these approaches, our approach more effectively learns a ``clean label'' multi-label classifier jointly with the noise distribution within the framework of the neural network. We do not assume the availability of any clean labels. Our model resembles a student-teacher model~\cite{Ba14,Hinton14,Paz16}, where a compact or compressed network (student model) is trained to reproduce the output of a deeper network (teacher model). In our case, we train a classifier network with a noise modeling network, where the noise modeling network (teacher) infers the posterior of the hidden true labels during training via backpropagation, as targets for multi-label classifier (student) training.

\section{Our Approach}
\subsection{Problem Formulation}

\newcommand{\image}{\mathcal{I}}
\newcommand{\tp}{\tilde{p}}
\newcommand{\zp}{\tp(z^c|\image)}
\newcommand{\qij}{q^c_{ij}}
\newcommand{\qijFI}{q^{c'}_{ij}}
\newcommand{\uc}{u^c_{ij}}
\newcommand{\bc}{b^c_{ij} }

\begin{figure*}[t]
  \begin{center}
    \includegraphics[width=0.95\linewidth]{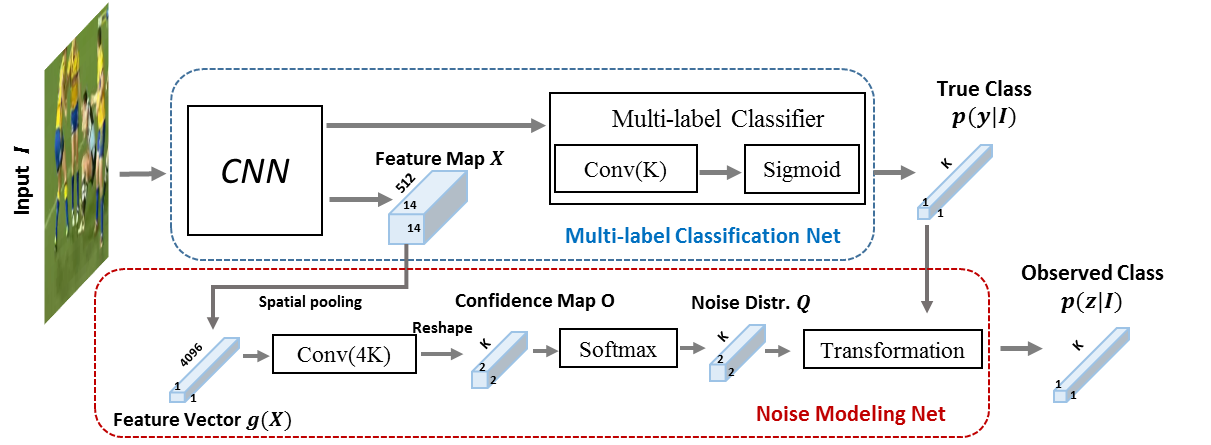}
  \end{center}
  \caption{Overall framework of our approach.~\emph{Conv(x)} is a convolution layer with x kernels of size 1x1. $K$ is the total number of classes.~\emph{Softmax} converts confidence map $O$ to noise distribution matrix $Q$. (Top) The multi-label classification net (MLCN) follows the structure of the fully convolutional VGG16 network (VGG16) in~\cite{Long15}.~\emph{CNN} is its sub-network including 15 convolution layers and 5 max-pooling layers, which outputs feature map $X$ to the noise modeling network (NMN). (Bottom) The proposed NMN learns the noise distribution Q, which is used to transform the predicted class distribution to the observed class distribution via the `Transformation' layer.}
  \label{overview2}
\end{figure*}

Our goal is to learn a multi-label classifier given a noisy labeled dataset. Let $\mathcal{I}$ denote an input image or video with observed noisy labels: $\mathbf{z}=[z^1,\dots,z^K]$, where $z^c\in\{0,1\}$.  $z^c$ is an indicator variable denoting whether the input $\mathcal{I}$ is tagged with label $c$. $z^c$ can be obtained from annotators, tags from its surrounding texts or keywords from a search engine. $K$ is the total number of possible labels. We denote another indicator variable $y^c\in\{0,1\}$ as the unobserved true label indicating whether the label $c$, in fact, belongs to $\mathcal{I}$.  Missing label noise for class $c$ occurs when $\image$ is labeled as not belonging to $c$ when in fact, it does (\textit{i.e.}, $z^c=0$, $y^c=1$ ). Incorrect label noise for class $c$ occurs when $\image$ is tagged with $c$  when in fact, it does not ($z^c=1$, $y^c=0$).
In the case of the noise-free model without the use of the NMN with the true labels $y^c$ observable, i.e. $y^c = z^c$, we can effectively train the MLCN by maximizing the log-likelihood, which takes the form of a cross entropy between provided labels $\mathbf{y}$ and the classifier predicted probabilities $f(\mathcal{I})=[\delta{(o^1)},\dots,\delta{(o^K)}]$:

\abovedisplayskip=1pt
\belowdisplayskip=1pt
\begin{align}
L(f(\mathcal{I}),\mathbf{y}) &= \sum_{c=1}^K{L(f(\mathcal{I}), y^c)}  \nonumber \\
&= \sum_{c=1}^{K}{y^c\log{\delta{(o^c)}}+(1-y^c)\log{(1-\delta{(o^c)})}} 
\label{crossentropy}
\end{align}
where $\delta{(a)}=1/(1+e^{-a})$ is the sigmoid function and $o^c$ is the logit (activation) computed for each class $c$. $\delta{(o^c)}$ is the estimated probability, i.e $\tp(y^c=1|\mathcal{I})=\delta{(o^c)}$, Note that in multi-class classification, a softmax would be used to estimate the joint probabilities across classes while in  multi-label classification, each label's probability is determined separately with sigmoid functions.

In the case of noisy annotations, one does not observe $y^c$ during the training process. If one simply assumes $y^c = z^c$ during classifier training, the training process will penalize models that correctly predict a low probability for samples affected by incorrect label noise. This can bias the final classifier to have a higher false positive rate. Similarly, missing label noises may result in a final classifier with higher miss rate. The Noise Modeling Net is needed to address both missing labels and incorrect labels.

\subsection{Dealing with Missing Labels and Incorrect Labels}

Motivated by the noise models in~\cite{Goldberger17,Bekker16} for the multi-class classification task, here we address missing labels and incorrect labels for multi-label classification by modeling a noise distribution $p(z^c=i|y^c=j,\mathcal{I})$ in multi-label images.

As shown in Figures~\ref{overview2}, our proposed approach augments the training of the {\it Multi-label Classification Net (MLCN)} which captures the conditional distribution $p(y^c|\mathcal{I})$ of the true label $y^c$ given image $\mathcal{I}$ , with the {\it Noise Modeling Net (NMN)} which captures the noise distribution $p(z^c|y^c,\image)$. To clearly differentiate between the proposed system with {\it NMN} and the {\it MLCN}, we refer to the full system with {\it NMN} as the {\it end-to-end} system.

Both the MLCN and the NMN are jointly optimized.  Denote $\tp(z^c|\image)$ as the estimated probability of the observed label $z^c$ and $\tp(\mathbf{z}|\image) = [\tp(z^1|\image)...\tp(z^K|\image)]$(we use $\tp$ to denote estimate probabilities and $\mathbf{z}$ for vectors).  The log-likelihood of the observed training data $\mathbf{z}$, for a single image $\image$, is the cross entropy given by

\begin{align}
L(\tp(\mathbf{z}|\image),\mathbf{z}) & = \sum_{c=1}^{K}{z^c\log{\zp}+(1-z^c)\log{(1-\zp)}}.
\label{obs-crossentropy}
\end{align}

$z^c$ and $y^c$ are connected by an unknown noise distribution $p(z^c|y^c, \mathcal{I})$, where the noisy label is dependent on both the true label and input features; it is the transition probability of noisy label $z^c$ from the true label $y^c$ given the input $\mathcal{I}$. The estimated feature-dependent noise transition $\qij$  can be defined as:

\begin{align}
\qij = \tp(z^c=i|y^c=j, \mathcal{I}) =& \frac{\exp((\uc)^Th(\mathcal{I})+\bc)}{\sum_i\exp((\uc)^T h(\mathcal{I})+\bc)}, \nonumber \\
~\textit{s.t.}~~i, j \in \{0,1\} 
\label{noisedistribution}
\end{align}
where $h(\mathcal{I})$ is the nonlinear function applied on input $\mathcal{I}$ and corresponds to the feature vector $g(X)$ in Figure~\ref{overview2}. The variables $\uc$ and $\bc$ are the parameters for class $c$ transition between $i$ and $j$. The activation $o^c_{ij}=(\uc)^Th(\mathcal{I})+\bc$ is an estimated unnormalized confidence score of transition from true label $j$ to noisy label $i$ for class $c$ (shown as {\it Confidence Map O} in Figure~\ref{overview2}).


In~\cite{Bekker16}, the noise distribution is assumed to only depend on the true label, independent of the input features. This feature independent transition, $\qijFI$ can be obtained by simplifying Equation~(\ref{noisedistribution}) to

\begin{align}
\qijFI = \tp(z^c=i|y^c=j) =& \frac{\exp(\bc)}{\sum_i\exp(\bc)}, ~\textit{s.t.}~~i, j \in \{0,1\}
\label{noisedistribution2}
\end{align}

For the rest of the paper, we will continue to use the feature dependent noise distribution with the understanding that it can be simplified to the feature independent case. Using the noise distribution, the estimated probability of observing a noisy label $z^c$ given input $\mathcal{I}$ is defined as:

\begin{align}
\tp(z^c=i|\mathcal{I}) &= \sum_{j} \qij \tp(y^c=j|\mathcal{I}), ~\textit{s.t.}~~i, j \in \{0,1\}
\label{observeddistribution}
\end{align}
which is the conditional distribution for observing $z^c = i$ while $\tp(y^c|\mathcal{I})$ is the estimated true class distribution. The estimated probability, $\tp(z^c=i|\mathcal{I})$, defined in Equation~(\ref{observeddistribution}), is optimized via Equation~(\ref{obs-crossentropy}) and connects the $\tp(y^c=j|\mathcal{I})$ estimated from the MLCN with the $\qij$ from the NMN.

From Equation~(\ref{observeddistribution}) and using Bayes rule, we can estimate, $\rho^c$, the posterior of the true label $y^c $ conditioned on $z^c$ by $\rho^{c} = \tp(y^c|z^c,\mathcal{I}) = \frac{\tp(z^c|y^c,\mathcal{I})\tp(y^c|\mathcal{I})}{\tp(z^c|\mathcal{I})}$.
Assuming $\rho^{c}_{ji} = \tp(y^c=j|z^c=i,\mathcal{I})$, let's discuss the properties of ideal MLCN and NMN when processing missing label and incorrect label training instances.

\vspace{-0.3cm}
\begin{itemize}
\item When encountering training samples with missing label noise (\textit{i.e.}, $z^c=0$ and $y^c=1$ ), we would be maximizing the log-likelihood of $\log{p(z^c=0|\mathcal{I})}$ and the ideal NMN should produce transition probability estimates of $q^c_{01} > q^c_{00}$  with the MLCN returning $\tp(y^c=1|\mathcal{I}) > \tp(y^c=0|\mathcal{I})$. The combination of $q^c_{01} > q^c_{00}$ and $\tp(y^c=1|\mathcal{I})  > \tp(y^c=0|\mathcal{I})$ will imply $\rho^c_{10} > \rho^c_{00}$.
\item Similarly, when training with instances of incorrect label noise, (that is, $z^c=1$ and $y^c=0$), a higher estimate of $q^c_{10} > q^c_{11}$ from NMN and MLCN with high confidence of $y^c=0$ will maximize the log-likelihood $\log{p(z^c=1|\mathcal{I})}$.
\end{itemize}
\vspace{-0.3cm}

\begin{figure*}[t]
\begin{center}
\centering
\subfigure[Missing labels]
{
\label{gradients:a}
\includegraphics[width=0.48\linewidth]{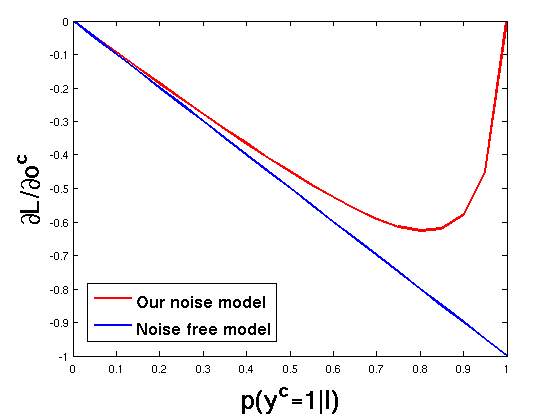}
}
\hspace{-0.4cm}
\subfigure[Incorrect labels]
{
\label{gradients:b}
\includegraphics[width=0.48\linewidth]{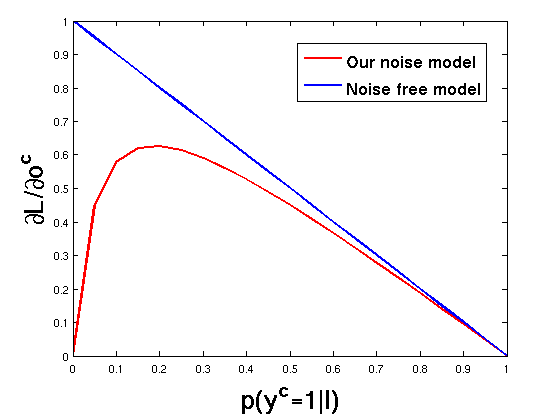}
}
\end{center}
\vspace{-0.2cm}
\caption{The derivative of training loss w.r.t. the learned classifier prediction for class $c$. The two probabilities for feature-dependent noise distribution are set to $q^c_{10}=q^c_{01}=0.05$. (a) Missing label scenario, i.e. $z^c=0$. (b) Incorrect label scenario with $z^c=1$. Compared with the noise free model, our model penalizes incorrect but confident predictions less. This property enables our model to address both missing label and incorrect label noises jointly.}
\label{gradients}
\vspace{-0.2cm}
\end{figure*}

These two properties allow the training of a good classifier from data with incorrect and missing label noises. Recall that the end-to-end system is trained by minimizing a cross-entropy loss (equivalent to maximizing the likelihood) between provided noisy labels and observed class distributions $\tp(z^c|\mathcal{I})$. Because our model factorizes over the classes $c$, and there is only a single latent variable $y^c$, the derivative of cross entropy loss $L$ with respect to the MLCN activation $o^c$ can be found directly. Let's examine how this derivative behaves under different noise conditions.

In the noise-free scenario with the noise-free model (\textit{i.e.}, $z^c=y^c$ and without NMN), the likelihood is described in Equation~(\ref{crossentropy}). We have $\frac{\partial{L}}{\partial{o^c}} \propto y^c-\tp(y^c=1|\mathcal{I})$.
Effectively, the learning procedure will try to make the prediction $\tp(y^c=1|\mathcal{I})$ as close to true label $y^c$ as possible. For the end-to-end model with the NMN, although the likelihood function in Equation~(\ref{obs-crossentropy}) is similar to Equation~(\ref{crossentropy}), the derivatives of the likelihood against the MLCN parameters have to back-propagate though the NMN.

\begin{itemize}
\item In the missing label scenario, the derivative $\frac{\partial{L}}{\partial{o^c}} \propto \rho^{c}_{10} - \tp(y^c=1|\mathcal{I})$. That is to say, the training makes the prediction $\tp(y^c=1|\mathcal{I})$ approximate the posterior probability of the true label $y^c=1$ given the noisy value $z^c=0$ and input $\mathcal{I}$. This will cause the classifier to be penalized less for making a confident but incorrect prediction.
\item In the incorrect label scenario, the derivative takes the form of $\tp(y^c=0|\mathcal{I})-\rho^c_{01}$. The training encourages the prediction $\tp(y^c=0|\mathcal{I})$ to approximate the posterior probability of the true label $y^c=0$ given the observed value $z^c=1$ and input $\mathcal{I}$.
\item For the correct label scenario, (\textit{i.e.}, $z^c=y^c=0$ and $z^c=y^c=1$), their derivatives take the form  $\tp(y^c=0|\mathcal{I})-\rho^c_{00}$ and  $\rho^c_{11}-\tp(y^c=1|\mathcal{I})$, respectively.
\end{itemize}

Figure~\ref{gradients} shows how the derivative changes with different values of predictions $\tp(y^c=1|\mathcal{I})$ using the noisy-free model and our noise model under a specific set of $q^c_{ij}$. When the model is consistent with the observed label, or $\tp(y^c = z^c |\mathcal{I})$ is high, the two models have similar derivatives.  However, when the model predicted label for the sample is very different from the observed label---a large $\tp(y^c = 1|\mathcal{I})$ for missing label noise and a small $\tp(y^c = 1|\mathcal{I})$ for incorrect label noise---the new model reduces the derivatives, reducing the impact of training under errorful labeling. By using the fact that $\sum_j \rho^c_{ji} = 1$, $\sum_j \tp(y^c=j|\mathcal{I}) = 1$ and some algebraic manipulation, we can show that the derivative has the form, $\frac{\partial{L}}{\partial{o^c}} \propto \rho^c - \tp(y^c=1|\mathcal{I})$.
That is to say, the proposed NMN generates the posterior probabilities of the true label $\rho^c$ as `soft target' labels to guide the MLCN training.

\subsection{Model Training and Noise Modeling Network}

Since hidden true class label can be considered as a random variable $y^c\in\{0,1\}$, as in~\cite{Goldberger17,Bekker16,Xiao15}, we could use the EM algorithm to learn the model parameters, where the E-step estimates the posterior of true labels and the M step updates the parameters of the noise distribution and classifier. However, the EM algorithm is known to get stuck in local optima easily, worse, if a large-scale neural network is used for classifier training, it may lead to slow optimization convergence and the network converging to a poor local minimum. Here we put all of these into an end-to-end deep learning framework and optimize them jointly. In fact, our network optimizes the same likelihood function optimized by EM-based algorithms. The proof is provided in the supplemental material.

We use a neural network to model the observed class distribution $\tp(z^c|\mathcal{I})$ in~(\ref{observeddistribution}). The overall framework is shown in Figure~\ref{overview2} and consists of two sub-networks: the \emph{MLCN} and the proposed \emph{NMN}. \emph{MLCN} has similar network structure as in fully convolutional VGG16 network~\cite{Long15} and includes two parts: \emph{CNN} and a \emph{Multi-label Classifier}. The \emph{CNN} includes 15 convolution layers and 5 max-pooling layers while the \emph{Multi-label Classifier} is modeled as a convolution layer with $K$ kernels of size $1\times1$ (`conv(K)' in Figure~\ref{overview2}) and a sigmoid layer. MLCN predicts the true class distribution $\tp(y^c|\mathcal{I})$ and feeds it to the NMN.

The NMN takes a feature map $X$ from the MLCN as inputs, and uses spatial pooling to obtain the input feature vector $g(X)$ (\textit{i.e.}, $h(\mathcal{I})$ in Equation~(\ref{noisedistribution})). Then it applies four linear classifiers for pairwise labels ($\uc$ and $\bc$) to the input feature vector, where the linear classifiers are modeled as a convolution layer with $4K$ filters of size $1\times1$ (`conv(4K)'), and followed by a reshape layer. The output of the reshape layer is a confidence map $O\in\mathbb{R}^{2\times2\times K}$. Next the confidence map is converted to the noise distribution matrix $Q$ by a softmax layer, which corresponds to the estimation of Equation~(\ref{noisedistribution}). Next $Q$ is fed into a \emph{Transformation} layer, which transforms the input true class probabilities $\tp(y^c|\mathcal{I})$ to the observed class probability $\tp(z^c|\mathcal{I})$ by computing a weighed sum of true class probabilities $\tp(y^c=0|\mathcal{I})$ and $\tp(y^c=1|\mathcal{I})$ with weights in noise distribution according to Equation~(\ref{observeddistribution}). For instance, we used $q^c_{10}$ and $q^c_{11}$ as weights to compute $\tp(z^c=1|\mathcal{I})$.  The whole network is a unified framework and is trained with the cross-entropy loss between noisy labels $z^c$ and $\tp(z^c|\mathcal{I})$. At test time, we remove the NMN and only use the MLCN to predict the true label $y^c$.

\subsection{Interaction with Multiple Instance Learning}

We also add the multiple instance learning (MIL) layer provided by~\cite{Fang15} to the MLCN. \emph{Image MIL} considers an input image as a bag $b_r$ and image regions as instances of the bag. The MIL layer predicts the probability $\tp^c_r$ of a bag containing concept $c$ based on the probabilities $\tp^c_{r,s}$ of individual instances (image regions) in the bag: $1-\prod_{s\in b_r}{(1-\tp^c_{r,s})}$. The MIL layer pools together the CNN features computed on the images regions spatially. We extend this approach to videos. We call it \emph{Video MIL}, where an input video is considered as a bag and each image region from each video frame is an instance in the bag. 

MIL can be viewed as an approach to compensating for label noise. Consider the task of training an image classifier from video frames. The individual frames are unlikely to be labeled, so they must be assumed from video-level labels. While a given label may apply to the video as a whole, its evidence may be missing from any given frame; this is analogous to the incorrect label case. With MIL, the label is applied to the bag of frames from the video instead of each individual frame, implicitly correcting for the incorrect label on individual frame.

Our NMN model handles the case where the label is incorrect for the entire bag. In addition, it can directly compensate for missing labels. The MIL layer also implicitly compensates for missing labels at the instance level, but has no mechanism for handling missing labels at the bag level. As we show in Section~\ref{sec:exp}, the two approaches are complementary and the NMN model still improves performance even when MIL layers are utilized.

\section{Experiments}
\label{sec:exp}
We evaluate our approach on an image captioning dataset: MSR-COCO~\cite{Lin14} and a video captioning dataset: MSR-VTT 2016~\cite{Xu16}. The MSR-COCO dataset includes $82,783$ training images and $40,504$ validation images. Each image has five human-annotated captions. Following~\cite{Fang15}, we equally split the validation set of MSR-COCO into validation and test sets. The MSR-VTT dataset provides $41.2$ hours of YouTube videos including $6,513$ training clips and $497$ validation clips. Each video clip comes with $20$ natural language descriptions.

Our approach is evaluated by image and video concept detection performance with missing label and incorrect label noises, and video captioning performances given the input visual features---which are the concept probabilities from our concept models. About concept vocabulary construction, for the MSR-COCO dataset, we used the concept vocabulary provided by~\cite{Fang15}, consisting of $1,000$ concepts extracted from captions of training images. These concepts belong to any part of speech including nouns, verbs, and adjectives. For the MSR-VTT video dataset, we constructed the concept vocabulary using the $1,087$ most common words in the captions of training videos.

To successfully train the network in Figure~\ref{overview2}, there are two stages: (1) we train the MLCN with cross-entropy loss on the noisy data directly;  (2) we add the proposed NMN to the MLCN and finetune them jointly from the MLCN model learned in stage one. Therefore one baseline model is performing classification using MLCN only.  To account for the additional epochs used for jointly fine tuning the NMN and MLCN, we created another baseline using the MLCN trained with the same number of additional training epochs used in fine tuning. Therefore we compare the following four approaches:

\begin{table*}[t]
\centering
\begin{tabular} {|c|c|c|c|c|c|c|c|c|}
\hline
& VB & NN & JJ & DT & PRP & IN & Others & All \\
\hline
count & 176 & 616 & 119 & 10 & 11 & 38 & 30 & 1000 \\
\hline
VGG16 & 18.01 & 34.94 & 20.33 & 32.8 & 19.2 & 21.83 & 16.16 & 28.97 \\
VGG16* & 18.1 & 35.14 & 20.45 & 32.86 & 19.32 & 21.93 & 16.31 & 29.13 \\
VGG16-NMN-FI  & 18.56 & 35.91 & 20.65 & \textbf{32.88} & \textbf{19.85} & 21.94 & \textbf{16.49} & 29.72 \\
VGG16-NMN-FD  & \textbf{18.78} & \textbf{36.43} & \textbf{21.14} & 32.82 & 19.62 & \textbf{22.13} & 16.45 & \textbf{30.14} \\
\hline
\hline
VGG16-MIL & 20.73 & 41.49 & 23.93 & 33.44 & 20.5 & 22.57 & 16.28 & 33.96 \\
VGG16-MIL~\cite{Fang15} & 20.7 & 41.4 & \textbf{24.9} & 32.4 & 19.1 & 22.8 & 16.3 & 34.0 \\
VGG16*-MIL & 20.87 & 41.78 & 24.08 & 33.57 & 20.65 & 22.77 & 16.47 & 34.2 \\
VGG16-MIL-NMN-FI  & 21.1 & 42.17 & 24.09 & 33.36 & 20.98 & 22.52 & 16.2 & 34.47 \\
VGG16-MIL-NMN-FD  & \textbf{21.59} & \textbf{42.78} &  24.71 & \textbf{33.74} & \textbf{21.35} & \textbf{23.12} & \textbf{16.62} & \textbf{35.04} \\
\hline
\end{tabular}
\vspace{5pt}
\caption{Concept detection performances on the MSR-COCO dataset in terms of average precision (\%) using different approaches. The second sub-table are the results from the model which is trained with the MLCN and MIL layer. The results of~\cite{Fang15} are copied from the original paper. The concepts belong to any part of speech (NN: Nouns, VB: Verbs, JJ: Adjectives, DT: Determiners, PRP: Pronouns, IN: Prepositions).}
\label{tb1}
\end{table*}

\begin{itemize}
\item \textbf{VGG16}: We train the MLCN on the noisy dataset directly, where the network is pretrained on ImageNet.
\item \textbf{VGG16*}: We train the VGG16 model with the same number of additional epochs as used in fine-tuning.
\item \textbf{VGG16-NMN-FI}: We add the proposed NMN to the MLCN without considering input features $h(\mathcal{I})$ nor learning convolution parameters $u_{ij}$ in Equation (\ref{noisedistribution}). Instead, the NMN learns the feature-independent noise distribution in Equation (\ref{noisedistribution2}). We fine tune the whole network with the model from VGG16 with the same number of additional epochs as in VGG16*.
\item \textbf{VGG16-NMN-FD}: Similar to VGG16-NMN-FI, we add NMN to the MLCN and train on top of the model from VGG16 with the same number of additional epochs as VGG16*, but the NMN is input feature dependent.
\end{itemize}

About the training parameters, we trained 3 epochs for VGG16 and 2 additional epochs for VGG16*, VGG16-NMN-FI and VGG16-NMN-FD. The same number of training epochs was also used when a multiple instance learning layer~\cite{Fang15} is added to the MLCN. They are denoted as VGG16-MIL, VGG16*-MIL, VGG16-MIL-NMN-FI and VGG16-MIL-NMN-FD. Models for both datasets are optimized with a learning rate of $1.5\times10^{-4}$. We used the feature map $X$ from the `pool5' layer of the VGG16 network as inputs to the NMN. During inference, we only use the trained MLCN to predict ``clean labels''.

\subsection{Detecting Concepts in Images with Missing Label Noise}

To evaluate the performance of our approach, we obtain ground-truth labels from each image's captions. Any word in any of the captions of an image which is in the concept vocabulary is included as a label for the image. Since these labels are from simple, human-annotated image captions, missing labels, on things such as small objects or object attributes, may occur. We measure the word detection performance in terms of average precision as in~\cite{Fang15}.

The results are summarized in Table~\ref{tb1}. We observe obvious performance improvements for nouns and adjectives using the proposed noise modeling network. As expected, the input feature-dependent NMN which considers the image content to determine noise distribution, achieved the best performance.

As shown in Table~\ref{tb1}, the MIL layer also improves the detection performance since better localization (\textit{i.e.}, scanning different image regions instead of single whole image) of possible objects can help to detect words as described in~\cite{Fang15}. Interestingly, our proposed NMN, which directly handles the missing label noises, can further improve performance, demonstrating the complementary nature of the two algorithms.

\subsection{Detecting Concepts in Images with Incorrect Label Noise}

\begin{table*}[t]
\centering
\begin{tabular} {|c|c|c|c|c|c|c|c|c|c|c|}
\hline
& & VB & NN & JJ & DT & PRP & IN & Others & All & All-MIL \\
\cline{2-11}
noise & count & 176 & 616 & 119 & 10 & 11 & 38 & 30 & 1000 & 1000 \\
\hline
\multirow{4}{*}{$20\%$}&VGG16 & 17.21 & 33.49 & 19.47 & 32.21 & 18.78 & 21.47 & 15.67 & 27.79 & 33.19 \\
&VGG16* & 17.29 & 33.63 & 19.58 & 32.27 & 18.94 & 21.56 & 15.78 & 27.92 & 33.41 \\
&VGG16-NMN-FI  & 17.76 & 34.57 & 19.88 & \textbf{32.49} & \textbf{19.33} & 21.55 & \textbf{15.89} & 28.62 & 33.68 \\
&VGG16-NMN-FD  & \textbf{18.14} & \textbf{35.26} & \textbf{20.48} & 32.34 & 19.13 & \textbf{21.82} & 15.82 & \textbf{29.19} & \textbf{34.41}\\
\hline
\hline
\multirow{4}{*}{$40\%$}&VGG16 & 16.33 & 31.88 & 18.36 & 31.7 & 18.02 & 20.81 & 15.16 & 26.46 & 32.21\\
&VGG16* & 16.44 & 32.08 & 18.48 & 31.75 & 18.19 & 20.90 & 15.21 & 26.62 & 32.38 \\
&VGG16-NMN-FI  & 16.63 & 32.88 & 18.78 & \textbf{32.09} & 18.37 & 20.85 & \textbf{15.38} & 27.19 & 32.58\\
&VGG16-NMN-FD  & \textbf{17.31} & \textbf{33.94} & \textbf{19.48} & 31.96 & \textbf{18.46} & \textbf{21.2} & 15.3 & \textbf{28.06} & \textbf{33.56}\\
\hline
\hline
\multirow{4}{*}{$60\%$}&VGG16 & 15.05 & 29.56 & 16.97 & 30.81 & 17.32 & 19.73 & 14.06 & 24.54 & 30.33\\
&VGG16* & 15.12 & 29.73 & 17.07 & 30.91 & 17.34 & 19.82 & 14.16 & 24.68 & 30.5\\
&VGG16-NMN-FI & 15.05 & 29.91 & 16.98 & \textbf{31.29} & 16.97 & 19.74 & 14.46 & 24.78 & 30.6\\
&VGG16-NMN-FD & \textbf{15.96} & \textbf{31.66} & \textbf{17.97} & 31.19 & \textbf{17.86} & \textbf{20.23} & \textbf{14.67} & \textbf{26.17} & \textbf{31.85} \\
\hline
\hline
\multirow{3}{*}{$80\%$}&VGG16 & 12.58 & 25.18 & 13.61 & 29.19 & 14.29 & 18.11 & 12.40 & 20.86 & 26.95 \\
&VGG16* & 12.65 & 25.3 & 13.72 & 29.30 & 14.40 & 18.22 & 12.45 & 20.96 & 27.1 \\
&VGG16-NMN-FI & 11.1 & 23.39 & 12.79 & \textbf{29.61} & 13.61 & 17.59 & 12.26 & 19.37 & 26.1 \\
&VGG16-NMN-FD & \textbf{13.48} & \textbf{27.19} & \textbf{14.71} & 29.44 & \textbf{15.27} & 18.62 & 12.32 & \textbf{22.41} & \textbf{28.5}\\
\hline
\end{tabular}
\vspace{5pt}
\caption{Concept detection performances on the MSR-COCO dataset in terms of average precision (\%) using different approaches. The four sub-tables correspond to the 20\%, 40\%, 60\% and 80\% incorrect label scenarios. The `ALL-MIL' column are the results from the model which is trained with the MLCN and MIL layer.}
\label{tb2}
\end{table*}

\begin{table*}
\centering
\begin{tabular} {|c|c|c|c|c|c|c|c|c|c|}
\hline
training & & VB & NN & JJ & DT & PRP & IN & Others & All \\
\cline{2-10}
level&count & 265 & 620 & 104 & 9 & 16 & 33 & 40 & 1087 \\
\hline
\hline
\multirow{4}{*}{Frame}&VGG16 & 28.82 & 37.21 & 25.79 & 45.3 & 31.87 & 42.49 & \textbf{28.63} & 33.9 \\
&VGG16* & 28.87 & 37.48 & 25.88 & 45.24 & \textbf{31.92} & 42.52 & 28.6 & 34.08 \\
&VGG16-NMN-FI & \textbf{29.81} & \textbf{38.34} & 26.19 & \textbf{45.32} & 31.49 & \textbf{43.04} & 27.7 & \textbf{34.8} \\
&VGG16-NMN-FD & 29.52 & 37.87 & \textbf{27.21} & 43.51 & 31.0 & 41.7 & 27.32 & 34.49 \\
\hline
\hline
\multirow{4}{*}{Video}&VGG16-MIL & 26.99 & 33.48 & 21.87 & 41.68 & 31.72 & 41.23 & 26.15 & 30.79 \\
&VGG16*-MIL & 26.97 & 33.48 & 21.9 & 41.7 & 31.68 & 41.25 & 26.21 & 30.8 \\
&VGG16-MIL-NMN-FI & 28.0 & 35.12 & 22.59 & \textbf{43.01} & 31.68 & \textbf{41.82} & 26.6 & 32.09 \\
&VGG16-MIL-NMN-FD & \textbf{28.15} & \textbf{36.36} & \textbf{24.9} & 42.83 & \textbf{32.94} & 41.44 & \textbf{27.08} & \textbf{33.08} \\
\hline
\end{tabular}
\vspace{5pt}
\caption{Concept detection performance on the validation set of the MSR-VTT dataset in terms of average precision (\%) using different approaches.}
\label{tb3}
\end{table*}

To demonstrate our approach's ability to handle incorrect label noises, we artificially generate incorrect labels for each training image. Note that there are five ground-truth captions for each image in this dataset. We can effectively generate incorrect labels by randomly replacing some of the reference captions by other captions randomly selected from other images. For example, our $20\%$ incorrect label noise case was generated in this fashion: we replace one of the five ground-truth captions by another caption randomly sampled from another image. Similarly, we generated $40\%$, $60\%$ and $80\%$ incorrect label noises for each image by sampling two, three and four captions from other images, respectively. Note that each ground-truth caption of an image may use unique words not used in the other ground-truth captions. Thus, replacing some of the ground-truth captions may also introduce some missing label noises.

The results are summarized in Table~\ref{tb2}. Our NMN shows consistent performance improvements over baseline methods (VGG16 and VGG16*) in three levels (20\%, 40\% and 60\%) of incorrect label noise. As shown in the right-most column (All-MIL) of Table~\ref{tb2}, we observe that the MIL layer helped to correct the incorrect labels during training since the VGG16 detection results are significantly improved over all different levels of noise. However, combining the NMN with the baseline model trained with the MIL layer further improved performance. VGG16-NMN-FI, while not as good as the VGG16-NMN-FD, is mostly better than the baseline VGG16 except at the 80\% level of incorrect label noise. In this extreme noise condition, feature dependent NMN is necessary.

The observed class distribution $\tp(z^c|\image)$ is considered as human-centric predictions in~\cite{Misra16} and works well in missing label scenarios based on their experiments. In order to see whether the observed class distribution works in the incorrect label scenarios, we test this observed class distribution under the 80\% incorrect label noise scenario. We got 21.98\% and 28.22\% mean average precision using the observed class distributions from our VGG16-NMN-FD model trained without MIL layer and with MIL layer, respectively. The results using observed class distribution became worse than using the estimated true class distribution $\tp(y^c|\image)$, which is consistent with our expectation.

\begin{figure*}[t]
  \begin{center}
    \includegraphics[width=0.8\linewidth]{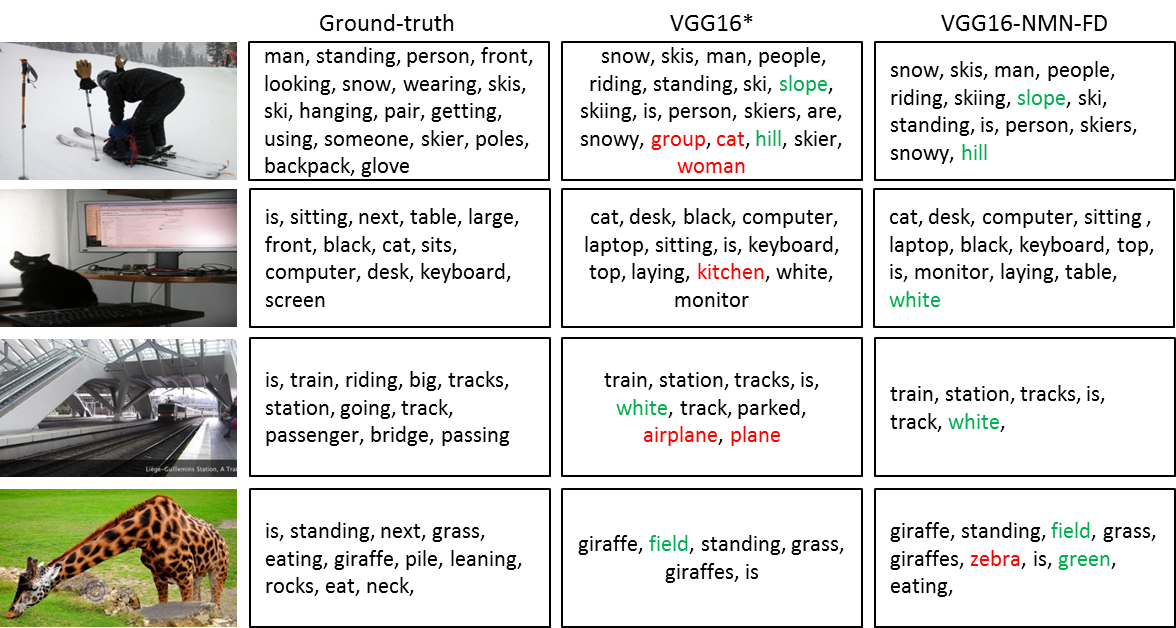}
  \end{center}
  \caption{Qualitative results of concept detection on the MSR-COCO dataset using different models trained under the 40\% incorrect label noise scenario. For visualization purpose, here we only show verbs, nouns and adjectives. The words in red are incorrect labels and the words in green are missing labels (\textit{i.e.} not mentioned in the ground-truth captions of images).}
  \label{example-coco}
\end{figure*}

Qualitative results are shown in Figure~\ref{example-coco} where these compared models are trained under 40\% incorrect label noises. Compared with baseline VGG16*, VGG16-NMN-FD can get better results in word detection since our model can model missing label and incorrect label noise.

\subsection{Detecting Concepts in Videos with Both Types of Noise}

\label{sec:videoconcept}
We believe incorrect label and missing label noises are common for video concept training. Similar to how we use captions to generate labels for images, we generate the concept labels for each video based on its $20$ video descriptions. The missing label noise is unavoidable since the labels are generated from annotated captions which do not fully describe all the details of the video. To train a concept classifier from videos, a simple and common approach is to extract video frames from each video, and then consider video-level labels as frame-level labels, such as in~\cite{Shen17}. This frame sampling and label assignment strategy will also generate incorrect label noise.

We uniformly extracted $26$ frames from each video, and these $26$-frame video representations are used for both training and testing. We tried two training settings: (1) \textbf{Frame-level Classifier}: we trained a frame-level concept classifier by simply considering video-level labels as frame-level labels. During testing, since the performance evaluation is based on video-level, we obtain a video-level prediction for each testing video by pooling the predictions from the $26$ frames with max-pooling. (2) \textbf{Video-level Classifier}: During training and testing, we aggregated the frame-level predictions across $26$ frames per video to obtain a video-level prediction by using the proposed \emph{video MIL} layer, an extension of the image MIL layer in~\cite{Fang15} originally developed for image classification. Here, a video is considered as a bag and all image regions of each video frame are its instances. The video MIL layer pools together the CNN features computed on instances temporally in addition to spatially.

For the frame-level classifier, we added the NMN to the MLCN (VGG16) without using the image MIL layer. For the video-level classifier, each video is considered as a mini-batch and fed into the MLCN, where a video MIL layer and NMN are added. We compared our NMN approach with other two baseline approaches in Table~\ref{tb3}. As shown in Table~\ref{tb3}, our approach can get better results in these two settings for verbs, nouns and adjectives, which often correspond to concrete regions in an image.

\subsection{Video Captioning Performance}
To show the effectiveness and representation power of concepts detected by our approach, we trained a video captioning model based on extracted concept features ($1,087$ concept probabilities). Our video captioning model is the sequence-to-sequence video-to-text (S2VT) model in~\cite{Venugopalan15}, which uses the Long Short Term Memory (LSTM) based encoder-decoder framework to map a sequence of video frames to a sequence of words. We used a vocabulary of $23,647$. The LSTM hidden state size is 700. We embed the input concept features and one-hot representations for caption words to a lower dimensional space, where the parameters of the embedding layers are jointly trained with the LSTM network parameters. The dimensions of the embedded representations for ground-truth caption words and concept features are $300$ and $700$, respectively.

We applied three trained concept models---VGG16*-MIL, VGG16-MIL-NMN-FI and VGG16-MIL-NMN-FD networks from Section~\ref{sec:videoconcept}---to extract concept features from each video frame as input features to the LSTM captioning network. As in~\cite{Ramanishka16,Shen17}, we report captioning performance on the validation set in terms of machine translation metrics, including BLEU-4, METEROR, CIDEr and ROUGE-L. In Table~\ref{tb4}, we can see that the concept features from the models trained with the NMN can achieve better captioning results compared with VGG16*-MIL, since the NMN enables the MLCN to predict correct and more related concepts from each frame, encouraging the language model to select appropriate input elements based on relevant concepts detected in the frame. Our results are better than state-of-the-art approaches including~\cite{Shen17,Venugopalan15a,Venugopalan15}, and comparable to C3D and ResNet features reported in~\cite{Ramanishka16}. Compared with~\cite{Yao15}, our method outperforms in all metrics except CIDEr.

\begin{table}
\centering
\begin{tabular} {|c|c|c|c|c|}
\hline
Model & CIDEr & Bleu-4 & ROUGE-L & METEOR  \\
\hline
VGG16*-MIL & 34.8 & 33.9 & 56.8 & 25.6  \\
VGG16-MIL-NMN-FI & 36.5 & 36.2 & 58.3 & 26.4 \\
VGG16-MIL-NMN-FD & 34.9 & 35.7 & 58.2 & 26.4 \\
\hline
MIMLL~\cite{Shen17} & 32.6 & 33.7 & 56.9 & 25.9 \\
Mean-Pooling~\cite{Venugopalan15a} & 35.0 & 30.4 & 52.0 & 23.7 \\
Soft-Attention~\cite{Yao15} & 37.1 & 28.5 & 53.3 & 25.0 \\
S2VT~\cite{Venugopalan15} & 35.2 & 31.4 & 55.9 & 25.7 \\
C3D~\cite{Ramanishka16} & 38.9 & 37.4 & 59.4 & 26.4 \\
ResNet~\cite{Ramanishka16} & 40.0 & 38.9 & 60.5 & 26.9 \\
\hline
\end{tabular}
\vspace{5pt}
\caption{Video captioning performances on the validation set of the MSR-VTT dataset using different approaches. The results of MIMLL, Mean-Pooling, Soft-Attention and S2VT are copied from~\cite{Shen17}, while the results of C3D and ResNet are from~\cite{Ramanishka16}.}
\label{tb4}
\end{table}

\section{Conclusion}
We introduced an auxiliary noise modeling network to tackle label noise in training of a multi-label classifier. In contrast to earlier work, our approach does not need any portion of data with clean labels to estimate the noise model; it is capable of handling both incorrect and missing labels. Unlike the EM-based approaches, it does not suffer from the tractability issues in the case of larger datasets because the classification and noise models are estimated simultaneously via back-propagation. We demonstrated consistent performance gains in experiments carried out using the MSR-COCO and the MSR-VTT datasets. We also demonstrated the complementarity of our approach with the multiple instance learning technique, employed by most of the state-of-the-art systems. Finally, we carried out video captioning experiments using the MSR-VTT datasets and showed that concept probabilities generated by the system trained with the auxiliary NMN serve as better input features to the sequence-to-sequence video-to-text model than concept probabilities generated by the baseline system. Our future work to extend the NMN to model the semantic relations between true labels, and explore how to learn these relations from noisy data directly.


{\small
\bibliographystyle{ieee}
\bibliography{eccv-NMN}
}

\newpage
\section*{Appendix}
\subsection*{Proof of deep neural networks with noise modeling network (NMN) optimizing the same likelihood function as the noise modeling approach by expectation-maximization (EM) algorithm}

In this supplementary material, we will describe the EM-based algorithm and end-to-end deep neural network based algorithm, which in fact optimize the same likelihood function. In the training phase, we are given only a noisy image dataset, where each image $\image$ has noisy labels $\mathbf{z}=[z^1,\dots,z^K]$, which are viewed noisy versions of hidden true labels $[y^1,\dots,y^K]$. $z^c\in\{0,1\}$ is an indicator variable denoting whether the input $\image$ is tagged with label $c$. $y^c\in\{0,1\}$ is another indicator variable indicating whether the label $c$, in fact, belongs to $\image$. $z^c$ and $y^c$ are connected by an unknown noise distribution $p(z^c|y^c, \image)$, which is the transition probability of noisy label $z^c$ from the true label $y^c$ given the input $\image$.

\subsection{EM algorithm}
Our goal is to train a multi-label classification network $p(y^c|\image; \theta_1)$, where $\theta_1$ is the network parameter set. We follow the EM learning approach described in~\cite{Bekker16}. In~\cite{Bekker16}, the noisy label $z^c$ is assumed to be independent of input features and only dependent on the true label $y^c$, \textit{i.e.}, $p(z^c|y^c)$. However, $z^c$ in our approach is dependent on input features in addition to the true label $y^c$, that is, $p(z^c|y^c, \image; \theta_2)$, where $\theta_2$ is the noise distribution parameter set. As pointed out in~\cite{Goldberger17}, there is no closed-form solution for $\theta_2$ since this noise distribution is input feature dependent, we can apply neural network to model this noise distribution and find the parameters. Therefore we could develop a neural network like our proposed noise modeling network to model noise distribution, and add it on top of the multi-label classification network.

Assuming $\theta=\theta_1\cup \theta_2$, the log-likelihood of model parameters is:
\setcounter{equation}{0}
\begin{align}
L(\theta) = \sum_{c=1}^K{z^c\log{p(z^c=1|\image;\theta)} +(1-z^c)\log{(1-p(z^c=1|\image;\theta))}}
\label{noiselabelcrossentropy}
\end{align}
where $p(z^c=1|\image;\theta) = \sum_{y^c\in\{0,1\}}p(z^c=1|y^c, \image; \theta_2)p(y^c|\image;\theta_1)$.

Based on the training data, the goal is to find classification network parameters $\theta_1$ and noise distribution parameters $\theta_2$ that maximize the likelihood function. Since true label $y^c\in\{0,1\}$ is hidden random variable, we can apply EM algorithm to find the maximum-likelihood parameters.

In the E-step, we estimate the posterior of hidden true label $y^c$ based on the noise label $z^c$ and \emph{current} classification network parameters $\theta_1^{(t)}$ and noise distribution parameters $\theta_2^{(t)}$:
\begin{align}
p(y^c|z^c,\image; \theta^{(t)}) = \frac{p(z^c|y^c,\image;\theta_2^{(t)})p(y^c|\image;\theta_1^{(t)})}{p(z^c|\image;\theta^{(t)})}
\label{posterior}
\end{align}

In the M-step, we need to optimize $\theta_1$ and $\theta_2$ given the estimated posterior probabilities of hidden true labels. For $\theta_1$, we need to maximize the following auxiliary function:
\begin{align}
Q(\theta_1)=\sum_{c=1}^{K}{\sum_{y^c\in\{0,1\}}p(y^c|z^c,\image; \theta^{(t)})\log{p(y^c|\image;\theta_1)}},
\end{align}
which can be written as:
\begin{align}
Q(\theta_1) &= \sum_{c=1}^{K}p(y^c=1|z^c,\image; \theta^{(t)})\log{p(y^c=1|\image; \theta_1)} \\
&+ (1 - p(y^c=1|z^c,\image; \theta^{(t)}))\log{(1-p(y^c=1|\image; \theta_1))} \nonumber
\end{align}

Recall $\delta{(o^c)}=p(y^c=1|\image; \theta_1)$, where $o^c$ is the logit computed for each class $c$ and $\delta{(o^c)}$ is the estimated probability, we can easily obtain the derivative of $Q$ with respect to $o^c$:
\begin{align}
\frac{\partial{Q(\theta_1)}}{\partial{o^c}}=p(y^c=1|z^c,\image; \theta^{(t)}) - p(y^c=1|\image; \theta_1)
\label{derivative_o_1}
\end{align}

The derivative of $Q(\theta_1)$ with respect to $\theta_1$ can be easily computed using backpropagation from $\frac{\partial{Q(\theta_1)}}{\partial{o^c}}$ we computed. Given the updated classification network parameters $\theta_1^{(t+1)}$, we need to maximize the likelihood function in Equation (\ref{noiselabelcrossentropy}) for $\theta_2$:
\begin{align}
L(\theta_2; \theta_1^{(t+1)}) = \sum_{c=1}^Kz^c\log{p(z^c=1|\image;\theta_2,\theta_1^{(t+1)})} \nonumber \\
	+(1-z^c)\log{(1-p(z^c=1|\image;\theta_2, \theta_1^{(t+1)}))}
\end{align}
where $p(z^c=1|\image;\theta_2, \theta_1^{(t+1)}) = \sum_{y^c\in\{0,1\}}p(z^c=1|y^c, \image; \theta_2)p(y^c|\image;\theta_1^{(t+1)})$. The parameters $\theta_2^{(t+1)}$ can be estimated by maximizing this likelihood function.

\subsection{End-to-end Deep Neural Network}

As described in our paper, we model the observed class distribution $p(z^c|\image;\theta)$ by an end-to-end deep neural network (DNN) system, which consists of two sub-networks: the \emph{Multi-label Classification Net (MLCN)} and the proposed \emph{Noise Modeling Net (NMN)}. MLCN captures the conditional distribution $p(y^c|\mathcal{I}; \theta_1)$ of the true label $y^c$ given image $\mathcal{I}$ , while NMN captures the noise distribution $p(z^c|y^c,\image; \theta_2)$. We can find the network parameters $\theta=\theta_1\cup \theta_2$ by minimizing the cross entropy between provided noisy labels $\mathbf{z}$ and the estimated probabilities $\tp(\mathbf{z}|\image; \theta) = [\tp(z^1|\image; \theta)...\tp(z^K|\image; \theta)]$:
\begin{align}
L(\theta) = - \sum_{c=1}^{K}{[z^c\log{\zp}+(1-z^c)\log{(1-\zp)}]}
\label{obs-crossentropy}
\end{align}

Obviously, the EM based algorithm and this deep neural network based algorithm are actually optimizing exactly the same function.

It is sufficient to compute the derivatives of $L(\theta)$ with respect to the logit $o^c$, because the derivative of $L(\theta)$ with respect to other neural network parameters can be easily computed using backpropagation from the estimated $\frac{\partial{L(\theta)}}{\partial{o^c}}$. We use the chain rule to compute the partial derivative of $L(\theta)$ with respect to the logit $o^c$:
\begin{align}
\label{derivative_o_2}
\frac{\partial{L(\theta)}}{\partial{o^c}}=&\frac{\partial{L(\theta)}}{\partial{\tp(z^c|\image;\theta)}}\frac{\partial{\tp(z^c|\image;\theta)}}{\partial{\tp(y^c|\image;\theta_1)}}\frac{\partial{\tp(y^c|\image;\theta_1)}}{\partial{o^c}},
\end{align}
where the partial derivative $\frac{\partial{L(\theta)}}{\partial{\tp(z^c|\image;\theta)}} = \frac{-1}{\tp(z^c|\image;\theta)}$, the partial derivative
\begin{align}
\frac{\partial{\tp(z^c|\image;\theta)}}{\partial{\tp(y^c|\image;\theta_1)}} = \left\{
    \begin{array}{c l}	
         \tp(z^c|y^c=1,\image;\theta_2) - \tp(z^c|y^c=0,\image;\theta_2)  &  \text{ if $y^c=1$}, \\
         \tp(z^c|y^c=0,\image;\theta_2) - \tp(z^c|y^c=1,\image;\theta_2)  &  \text{ if $y^c=0$},
    \end{array}\right.
\end{align}
and $\frac{\partial{\tp(y^c|\image;\theta_1)}}{\partial{o^c}} = \left\{
    \begin{array}{c l}	
        \tp(y^c=0|\image;\theta_1)\tp(y^c=1|\image;\theta_1) & \text{ if $y^c=1$}, \\
        -\tp(y^c=0|\image;\theta_1)\tp(y^c=1|\image;\theta_1) & \text{ if $y^c=0$},
    \end{array}\right.$

Therefore equation~(\ref{derivative_o_2}) can be rewritten as:

\begin{align}
\label{derivative_o_2_1}
\frac{\partial{L(\theta)}}{\partial{o^c}} =& \frac{1}{\tp(z^c|\image;\theta)}(\tp(z^c|y^c=0,\image;\theta_2) - \tp(z^c|y^c=1,\image;\theta_2))  \\
  &\times\tp(y^c=0|\image;\theta_1)\tp(y^c=1|\image;\theta_1) \nonumber
\end{align}

By the definition of posterior of true label in equation~(\ref{posterior}), equation~(\ref{derivative_o_2_1}) becomes:

\begin{align}
\frac{\partial{L(\theta)}}{\partial{o^c}}=\tp(y^c=0|z^c,\image;\theta)\tp(y^c=1|\image;\theta_1) \nonumber \\
	- \tp(y^c=1|z^c,\image;\theta)\tp(y^c=0|\image;\theta_1)
\label{derivative_o_3}
\end{align}

By using the fact that $\sum_{y^c\in\{0,1\}}\tp(y^c|z^c,\image;\theta)=1$ and $\sum_{y^c\in\{0,1\}}\tp(y^c|\image;\theta_1)=1$, it becomes:
\begin{align}
\frac{\partial{L(\theta)}}{\partial{o^c}} =  \tp(y^c=1|\image;\theta_1) - \tp(y^c=1|z^c,\image;\theta)
\label{derivative_o_4}
\end{align}

We can interpret the derivative with respect to logit $o^c$ as encouraging MLCN to approximate the posterior of true label during the training, which is the same as in equation~(\ref{derivative_o_1}) and demonstrates that the two optimization problems are the same. Our main point in this section is that for our end-to-end deep neural network system, the MLCN model parameters $\theta_1$ and NMN model parameters $\theta_2$ are estimated via stochastic gradient descent during neural network training. We conclude that our end-to-end deep neural network and the EM algorithm are actually optimizing the same function, with the DNN training providing a scalable, practical solution to multi-label learning problems from noisy label datasets.

\end{document}